\documentclass[11pt, draftclsnofoot, onecolumn, journal]{IEEEtran}
\usepackage{cite}

\usepackage{color}

\ifCLASSINFOpdf
  \usepackage[pdftex]{graphicx}
\else
\fi
\usepackage[cmex10]{amsmath}
\usepackage{amssymb}

\usepackage{array}
\usepackage{float}

\usepackage{color}

\begin{document}
%
\title{Unsupervised Despeckling}
%
%
%

\author{Deepak~Mishra,~\IEEEmembership{}
        Santanu~Chaudhury,~\IEEEmembership{}
        Mukul~Sarkar~\IEEEmembership{}
        and~Arvinder~Singh~Soin~\IEEEmembership{}

\thanks{Manuscript received X, XXXX.}
\thanks{D. Mishra, S. Chaudhury and M. Sarkar are with the Electrical 
Engineering department, Indian Institute of Technology Delhi, 110016 New Delhi, 
India. (e-mail: deemishra21@gmail.com, santanuc@ee.iitd.ac.in, msarkar@ee.iitd.ac.in).
\newline A. S. Soin is with the Medanta Hospital, Gurgaon, India. (e-mail: absoin@gmail.com)}}%
%
%

\markboth{Journal,~Vol.~XX, No.~X, ~2017}%
{Shell \MakeLowercase{\textit{et al.}}: Bare Demo of IEEEtran.cls for Journals}
%



\maketitle

\begin{abstract}
Contrast and quality of ultrasound images are adversely affected by the excessive presence of speckle. However, being an inherent imaging property, speckle helps in tissue characterization and tracking. Thus, despeckling of the ultrasound images requires the reduction of speckle extent without any oversmoothing. In this letter, we aim to address the despeckling problem using an unsupervised deep adversarial approach. A despeckling residual neural network (DRNN) is trained with an adversarial loss imposed by a discriminator. The discriminator tries to differentiate between the despeckled images generated by the DRNN and the set of high-quality images. Further to prevent the developed DRNN from oversmoothing, a structural loss term is used along with the adversarial loss. Experimental evaluations show that the proposed DRNN is able to outperform the state-of-the-art despeckling approaches.

\end{abstract}

\begin{IEEEkeywords}
Ultrasound image, speckle, deep learning, GAN.
\end{IEEEkeywords}

%
\IEEEpeerreviewmaketitle

\section{Introduction}
%
%
%
%
\IEEEPARstart{S}{peckle} extent or size of speckle in ultrasound (US) images is a function of axial and lateral resolution of ultrasonic transducers. The transducers with small apertures are convenient to use, however, produce images with large speckle extent due to their poor lateral resolution~\cite{shung1992ultrasonic}. The reduction of speckle extent in such images is required to improve the image contrast and diagnostic quality \cite{mishra2017edge}. However, an excessive speckle filtering results in loss of the features presented by speckle characteristics \cite{ramos2015anisotropic, mishra2017edgeprob}. Thus, despeckling of US images is basically the reduction of speckle extent without losing the characteristics details of tissues.

The conventional approaches consider speckle as noise and try to reduce inter-pixel intensity variations \cite{tay2010ultrasound, lee1981speckle, kuan1987adaptive, yu2002speckle, aja2006estimation, yue2006nonlinear, krissian2007oriented, zhang2007nonlinear, coupe2009nonlocal, balocco2010srbf, deng2011speckle}. Most of these approaches suffer from oversmoothing and generate images with artificial appearances. Few recent approaches, for example \cite{mishra2017edgeprob, ramos2015anisotropic, munteanu2008speckle}, prohibit the filtering in high echogenicity regions to reduce the oversmoothing effect. However, they end up with no despeckling in the images of the organs containing large volumes of parenchymal tissues like liver. Further, these approaches require a precise tuning of the implementation parameters for every input image to obtain the desired results.
Speckle reduction has also been modeled as an optimization problem in  \cite{aysal2007rayleigh, aubert2008variational, bioucas2010multiplicative}. However, these approaches result in suboptimal performances due to oversimplified assumptions for example the speckle is assumed to have multiplicative nature, which is not entirely correct for the US images \cite{michailovich2006despeckling}.

Considering despeckling as an optimization problem, it can be solved using supervised and unsupervised learning. The supervised learning approaches appear to be impractical due to the unavailability of the required annotated training data.
Hence, in this work, a deep unsupervised learning method is proposed to reduce speckle extent while preserving the characteristic details. A deep residual network (ResNet) \cite{he2016deep}, as a part of a generative adversarial network (GAN), is trained to produce the despeckled images. 
GANs have recently been used in several computer vision applications for natural images \cite{goodfellow2014generative, radford2015unsupervised, ledig2016photo, wang2016generative, zhu2017unpaired, Shrivastava_2017_CVPR}. However, their use in US images is been limited to simulated data generation \cite{hu2017freehand}. In this work, GAN is used for US image despeckling and to best of our knowledge, this is the first attempt to use GAN for the processing of US images. Concurrent to this work, a GAN with Wasserstein distance and perceptual similarity for CT image denoising is proposed in \cite{yang2017low}, however the contrasting nature of the imaging modalities differentiates the two works.

The unsupervised training of the developed despeckling residual neural network (DRNN) is accomplished in two steps. The DRNN, which is used as the generator in GAN framework, is first trained to reproduce the inputs. Later, it is combined with a discriminator network and trained for the despeckling. A convolutional neural network (CNN) is used as the discriminator. Two sets of liver US images are created for GAN training. One set contains low-quality US images with high speckle extent whereas the second set contains the images with high contrast and considerably less speckle. During training, the generator network learns to reduce the speckle extent of the poor quality  images and produce similar speckle characteristics as the better quality set. The generator is trained with the adversarial loss imposed by the discriminator. However, to prevent the generator from copying the data from the high-quality set and maintaining the characteristic details of the input, the adversarial loss is combined with a structural loss. The structural loss works as a regularizer and helps in producing the desired outputs. 


The paper is organized as follows. Section II provides the details of the proposed method. Section III presents the experimental results which are compared with the existing approaches. Finally, conclusions are presented in section IV.

\section{Method}

\begin{figure*}[!t]
\centering
\begin{minipage}[]{0.9\linewidth}
  \centerline{\includegraphics[width=7in]{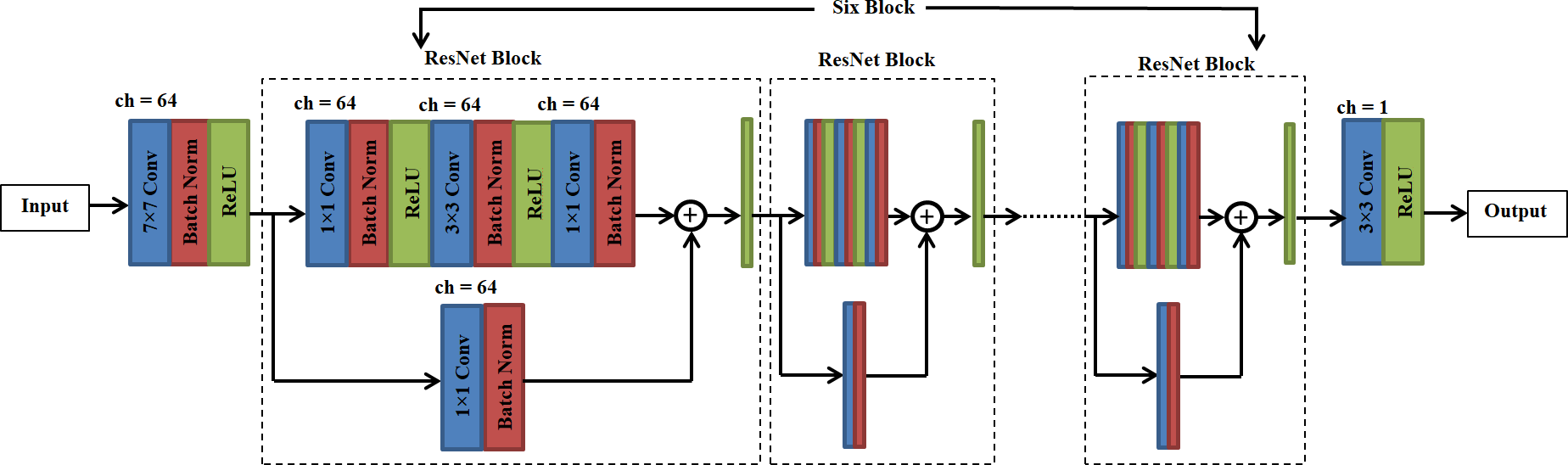}}
\end{minipage}
\caption[]{Despeckling or generator network architecture. The first convolutional (Conv) layer contains 7$\times$7 kernels and has 64 channels (ch). It is combined with batch normalization (Batch Norm) and rectified linear units (ReLU). The first layer is followed by a chain of six ResNet blocks and a final Conv layer.}
\label{G1}
\end{figure*}
\begin{figure*}[!t]
\centering
\begin{minipage}[]{0.9\linewidth}
  \centerline{\includegraphics[width=7in]{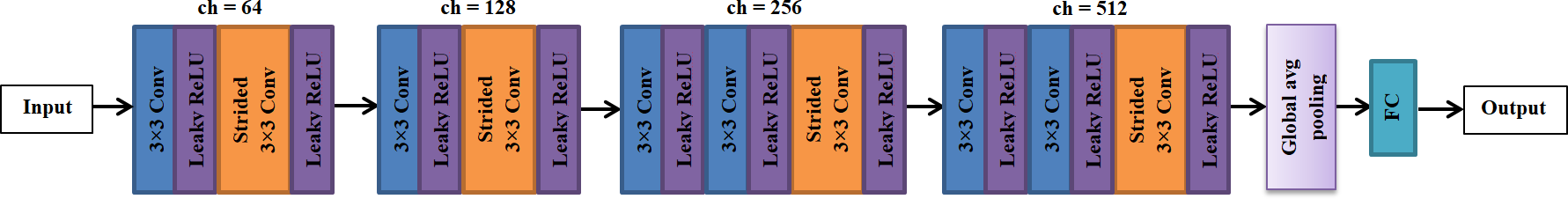}}
\end{minipage}
\caption[]{Discriminator network architecture. The strided convolutional layers are used to avoid the maxpooling layers along with the Leaky ReLU $(\alpha=0.2)$. The final output is a single value from the sigmoid activation.}
\label{D1}
\end{figure*}

In this work, US image despeckling is modeled as a single image input-output problem where the objective is to get a high-quality low speckle extent image ($\hat{x}$) given the speckle contaminated low-quality US image ($x$). GAN architecture used to accomplish the desired objective is explained below.

\subsection{Network Architecture}

GANs combine generative and discriminative models to implicitly learn the distribution of output class. The architecture of the generator ($\mathbb{G}$) used in this work is shown in Fig.~\ref{G1}. It contains a convolutional layer with batch normalization and rectified linear non-linearities (ReLU), followed by a chain of six ResNet blocks \cite{he2016deep} and a final convolutional layer for the desired output. As shown in Fig.~\ref{G1}, each ResNet block contains three convolutional layers and weighted shortcut connection between input and output of the block. Each convolutional layer, except the final layer, has 64 channels and is combined with batch normalization and ReLU non-linearities.

Similarly, the architecture of the discriminator ($\mathbb{D}$) is shown in Fig.~\ref{D1}. The findings of \cite{radford2015unsupervised} are adopted to build this network. All convolutional layers contain 3$\times$3 kernels and are combined with leaky ReLU ($\alpha=0.2$) \cite{maas2013rectifier, xu2015empirical} non-linearities. Further, strided convolutions, with stride of 2, are used to avoid the pooling layers. The number of channels is doubled after the strided convolution to have a uniform computational complexity of each convolutional layer. A global average pooling layer \cite{lin2013network} is used after the final convolutional layer to have a flexible input size and avoid overfitting associated with the conventional fully connected layers with a large number of neurons. At the end, a single neuron fully connected layer with sigmoid activation is used to get the desired output. 

\subsection{Training}
The generator $ \mathbb{G} $ is first trained to reproduce the input using $\ell_1$ loss as: 
\begin{equation}
L_1(\theta) = \mathbf{E}_x\big[\vert\vert\mathbb{G}(x;\theta)-x\vert\vert_1\big]
\label{l1_loss}
\end{equation}
where $\theta$ represent the parameters of $ \mathbb{G} $, $\vert\vert.\vert\vert$ is the $\ell_1$ norm and $\mathbf{E}[.]$ represents the expectation. $ \mathbb{G} $ is trained for 100 epochs on the samples of size 192$\times$192 obtained by randomly cropping the low-quality speckled images. Adam optimizer \cite{kingma2014adam} with an initial learning rate of $10^{-4}$ is used to optimize the parameters $\theta$. The batchsize used for training is 16.

Similarly, $ \mathbb{D} $ is trained to distinguish between the samples drawn from the set of high-quality US images ($y$) and the despeckled images $\left(\mathbb{G}(x;\theta)\right)$ generated from the set of low-quality speckled US images ($x$). The binary cross-entropy loss used in the training is:
\begin{equation}
L_\mathbb{D}(\phi) = \mathbf{E}_y\big[-\log\left(\mathbb{D}(y;\phi)\right)\big] + \mathbf{E}_x\big[-\log\left(1-\mathbb{D}\left(\mathbb{G}(x;\theta);\phi\right)\right)\big] 
\label{Dloss}
\end{equation}
where $\phi$ represents the parameters of $ \mathbb{D} $, which is trained for 5 epochs on the samples of size 128$\times$128. The training samples are randomly cropped in equal proportion from the high-quality US images and the despeckled images generated by $\mathbb{G}$. As the output of $ \mathbb{D} $ is a value resulting in from the sigmoid activation, the target labels for $y$ and $\left(\mathbb{G}(x;\theta)\right)$ are 1 and 0, respectively. $ \mathbb{D} $ is trained using Adam optimizer with an initial learning rate of $10^{-6}$ and batchsize 80.

The initial training of $ \mathbb{G} $ and $ \mathbb{D} $ is followed by the training of the two networks in GAN framework. Same input sizes as the initial training, but half of the batchsizes, are used due to the memory limitation of the available hardware (Nvidia GeForce GTX 1080 GPU, 8GB). $ \mathbb{D} $ is trained with the same loss as (\ref{Dloss}), however, $ \mathbb{G} $ is trained with a combination of adversarial and structural loss given by (\ref{Gloss}).
\begin{equation}
L_\mathbb{G}(\theta) = \mathbf{E}_x\big[-\log\left(\mathbb{D}\left(\mathbb{G}(x;\theta);\phi\right)\right)\big] + \lambda\left(L_1(\theta) + L_\text{MS-SSIM}(\theta)\right)
\label{Gloss}
\end{equation}
First term in (\ref{Gloss}) is the adversarial loss. In presence of this loss, the parameters $\theta$ are optimized such that the generated images are close to the high-quality US images. Further, the minimization of adversarial loss helps $ \mathbb{G} $ to implicitly learn the features used by $ \mathbb{D} $ to distinguish the high-quality US images and the generated despeckled images.  However, to avoid the overestimation of the features, which may result in loss of input image characteristics, the structural loss, downweighted with the parameter $\lambda (= 0.5)$, is used. Recent works on despeckling \cite{ramos2015anisotropic, mishra2017edgeprob} show that the structural similarity index measure (SSIM) \cite{wang2004image} is an important metric to evaluate the effect of overfilteirng. Further, as reported in \cite{zhao2017loss}, a combination of $\ell_1$ and multi-scale SSIM (MS-SSIM) loss is useful to obtain perceptually correct images. Thus, the structural loss is defined as the sum of $L_1(\theta)$ and $L_\text{MS-SSIM}(\theta)$ losses, where $L_1(\theta)$ is given by (\ref{l1_loss}) and $L_\text{MS-SSIM}(\theta)$ is defined as \cite{zhao2017loss}:
\begin{equation}
L_\text{MS-SSIM}(\theta) = 1-\mathbf{E}_x\left[l_M(\theta).\prod_{j=1}^{M}cs_j(\theta)\right]
\end{equation}
\begin{equation}
l_M(\theta) = \frac{2\mu_x\mu_{\hat{x}}+C_1}{\mu_x^2+ \mu_{\hat{x}}^2 + C_1}
\end{equation}
\begin{equation}
cs_j(\theta) = \frac{2\sigma^j_{x{\hat{x}}}+C_2}{{\sigma^j_x}^2+ {\sigma^j_{\hat{x}}}^2 + C_2}
\end{equation}
\begin{equation}
\mu_x = G_\sigma * x
\end{equation}
\begin{equation}
{\sigma^j_x}^2 = G_{\sigma^j} * x^2 - \mu_x^2
\end{equation}
\begin{equation}
\sigma^j_{x{\hat{x}}} = G_{\sigma^j} * (x.\hat{x}) - \mu_x\mu_{\hat{x}}
\end{equation}
where $\hat{x} = \mathbb{G}(x;\theta)$ is the despeckled image generated by $\mathbb{G}$. $G_{\sigma^j}$ is the Gaussian kernel used to calculate mean and variances of the input and output images. $\sigma^j$ is the standard deviation of $G_{\sigma^j}$ where $j\in\{1,2,...,M\}$ represents the size or scale of the kernel. ``$*$'' and ``$.$'' represent convolution and element-wise multiplication, respectively. The small constants $C_1$ and $C_2$ are added for stability. In this work $C_1$ and $C_2$ are chosen to be 0.01.

GAN is trained for 500 iterations. The parameters $\theta$ and $\phi$ are optimized by alternatively minimizing $L_\mathbb{G}(\theta)$ and $ L_\mathbb{D}(\phi) $. In a single iteration of GAN, $ \mathbb{D} $ is trained once whereas $ \mathbb{G} $ is trained for 20 times. The Adam optimizer with an initial learning rate of $10^{-6}$ for $ \mathbb{D} $ and an order less for $ \mathbb{G} $ is used. In each iteration, 50\% of samples in a minibatch used by $ \mathbb{D} $ are randomly selected from the high-quality set and remaining 50\% are generated by $ \mathbb{G} $ from the high speckle extent low-quality images. However, instead of generating all the 50\% samples using the current $ \mathbb{G}(\theta) $, 25\% samples are obtained from a buffer of the despeckled samples generated during the previous iterations. The buffer is updated after every iteration by replacing the old despeckled samples with the current ones, equal to the 25\% of the minibatch size. As reported in\cite{Shrivastava_2017_CVPR}, the buffering helps in the stabilization of the GAN training. The networks are is build on keras \cite{chollet2015keras}.

\section{Experiments and Results}

\subsection{Data}
\begin{figure}[!t]
\centering
\begin{minipage}[]{0.45\linewidth}
  \centerline{\includegraphics[width=3.8cm]{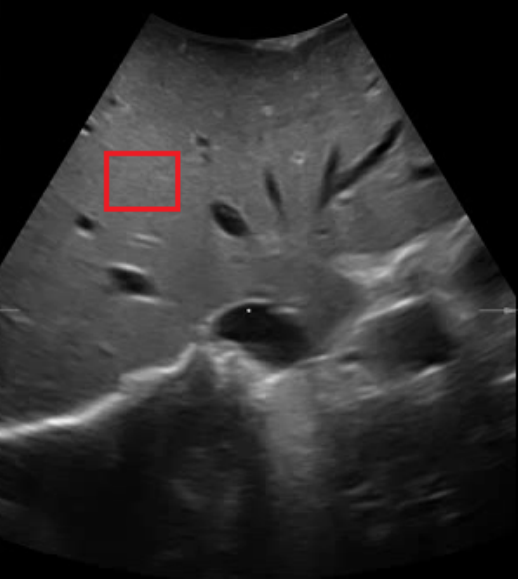}}
  \centerline{\footnotesize{(a) Image from RAB6-D scanner} }
\end{minipage}
%
\begin{minipage}[]{.45\linewidth}
  \centerline{\includegraphics[width=3.8cm]{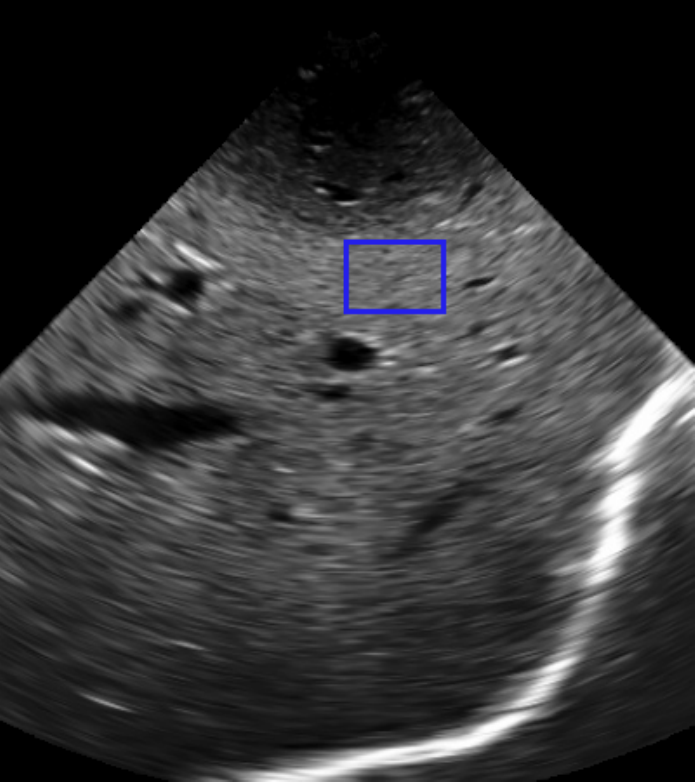}}
  \centerline{\footnotesize(b) Image from X7-2t scanner}
\end{minipage}
\caption[]{ Examples of (a) low speckle extent image ($y$) acquired using RAB6-D probe, (b) high speckle extent image ($x$) acquired using X7-2t probe. 
}
\label{sample}
\end{figure}

The RAB6-D convex array transducer with larger side of the aperture as 63.6 mm from GE Healthcare is used to acquire US images with low speckle extent. On the other hand, the US images with high speckle extent are acquired using the Philips X7-2t sector array probe with considerably smaller aperture and transducer elements. Acquisition is done in accordance with the regulations prescribed for medical experiments and with the full consent of the subjects. Sample images from the two probes are shown in Fig~\ref{sample}. Both the US images in Fig~\ref{sample} show liver of healthy subjects. However, the high speckle extent in the image from X7-2t probe affects the tissue contrast and detectability. Given the smaller aperture, the reduction in speckle extent can increase the utility of the probe for applications like intra-operative navigation and control.

The initial trainings of $ \mathbb{G} $ and $ \mathbb{D} $
are performed using 12,000 and 4,000 samples extracted from the images acquired using the two probes. For GAN training, all the images acquired using the two probes are cropped to the respective input sizes using a moving window with 50\% overlap. The total number of low-quality ($x$) and high-quality ($y$) samples available for training are 24,620 and 9,613, respectively. 

\begin{figure*}[!t]
\centering
\hspace{10pt}
\begin{minipage}[]{0.23\linewidth}
  \centerline{\includegraphics[width=3.8cm]{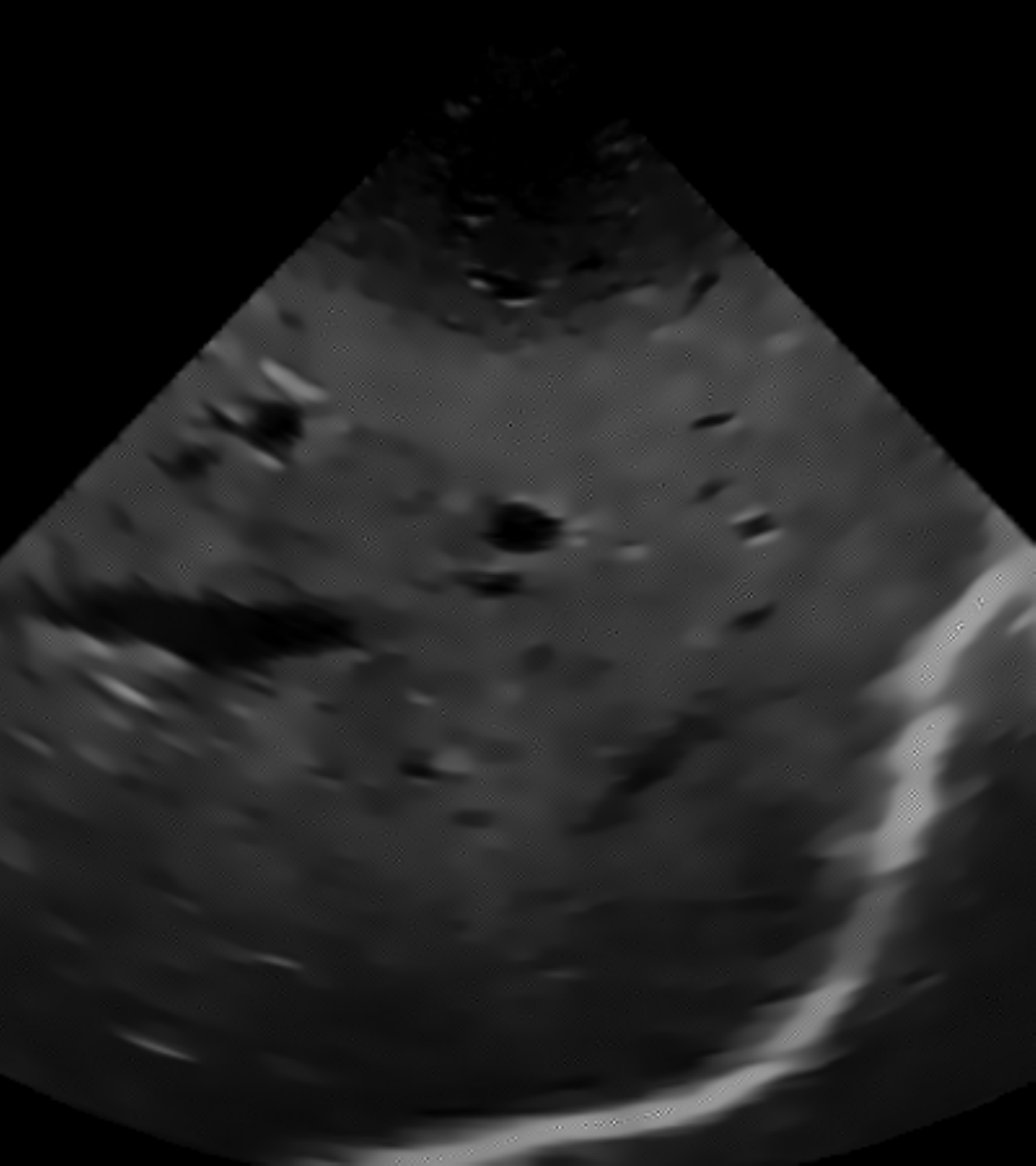}}
\end{minipage}
%
\begin{minipage}[]{.23\linewidth}
  \centerline{\includegraphics[width=3.8cm]{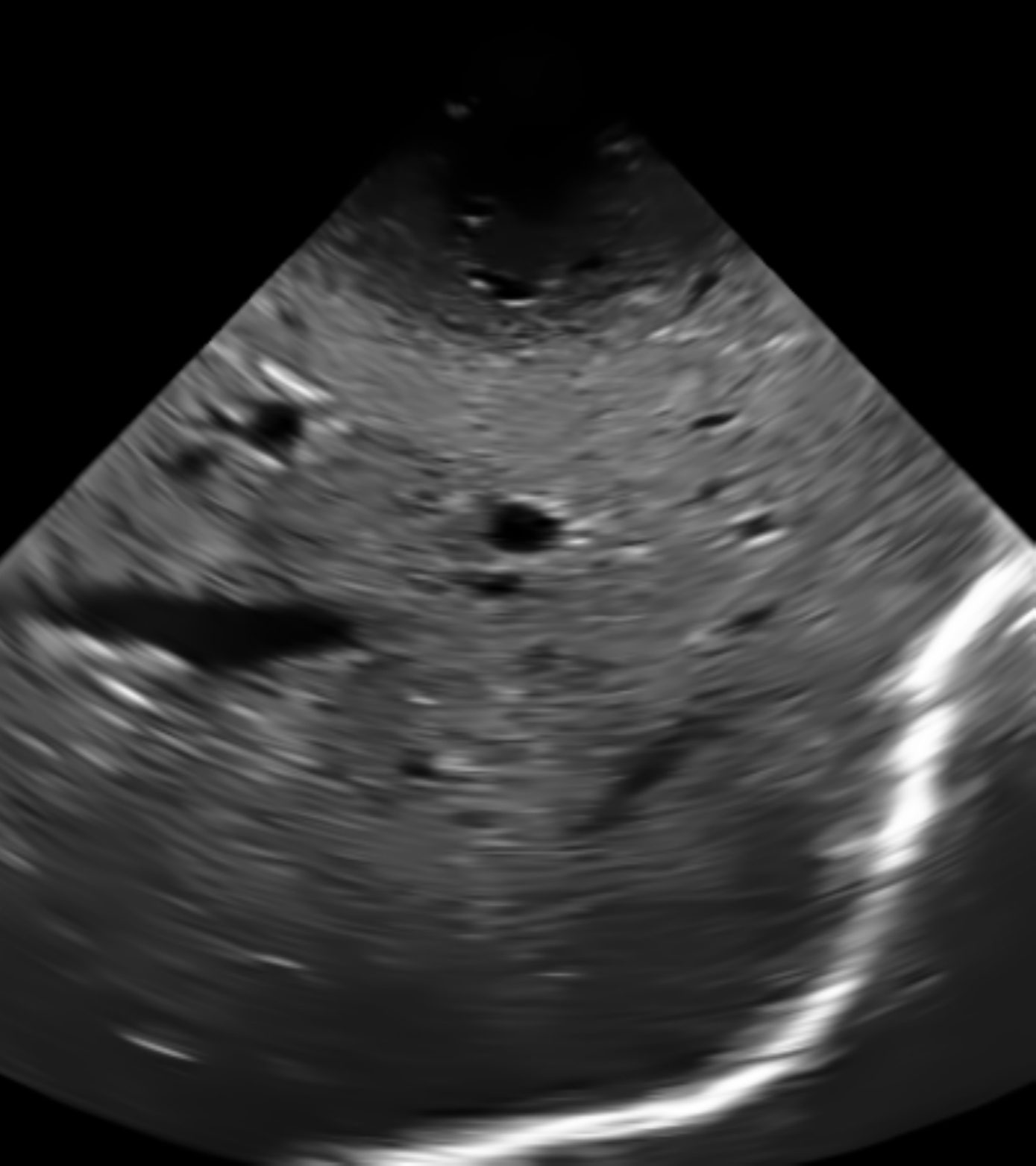}}
\end{minipage}
\begin{minipage}[]{0.23\linewidth}
  \centerline{\includegraphics[width=3.8cm]{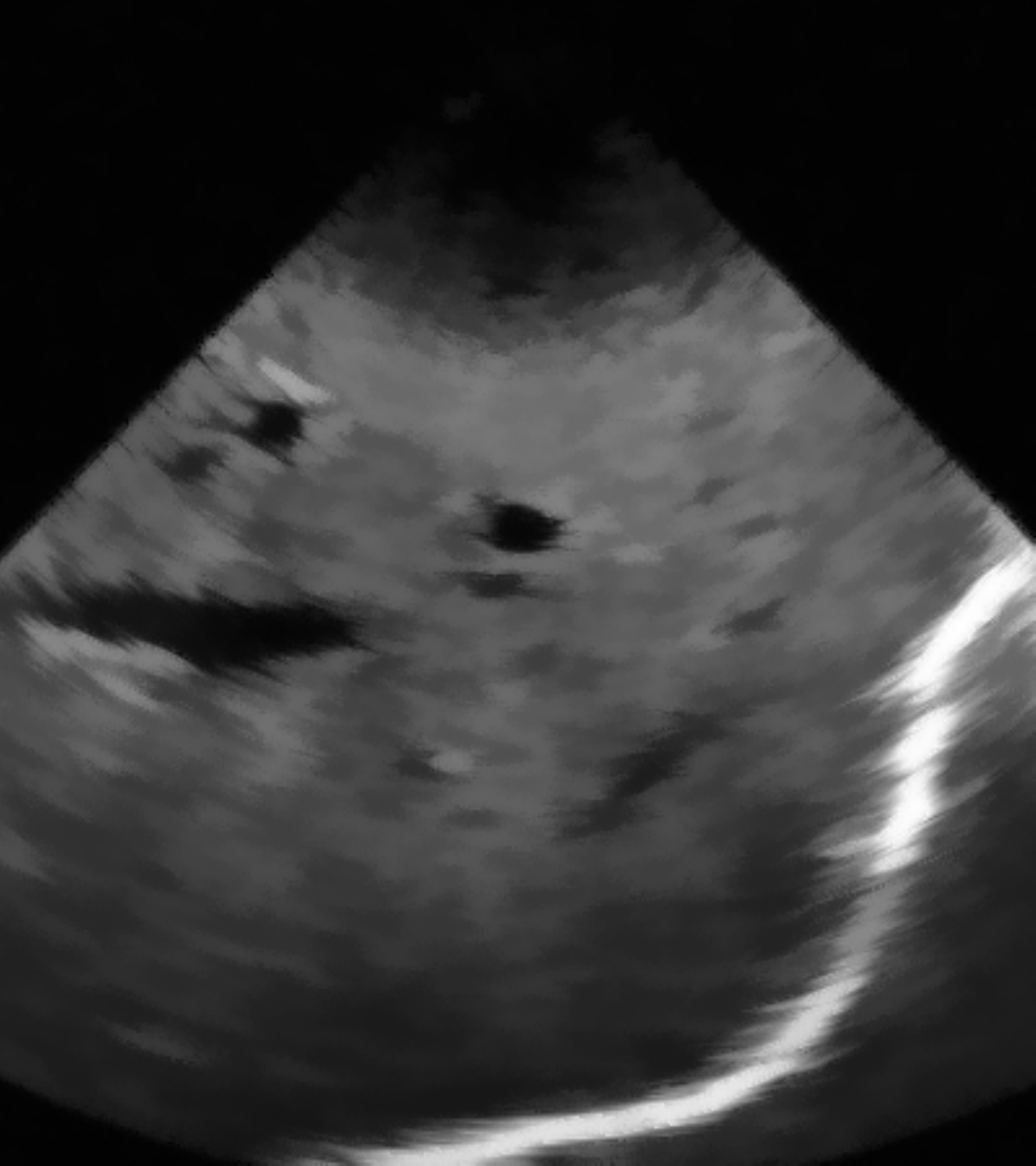}}
\end{minipage}
%
\begin{minipage}[]{.23\linewidth}
  \centerline{\includegraphics[width=3.8cm]{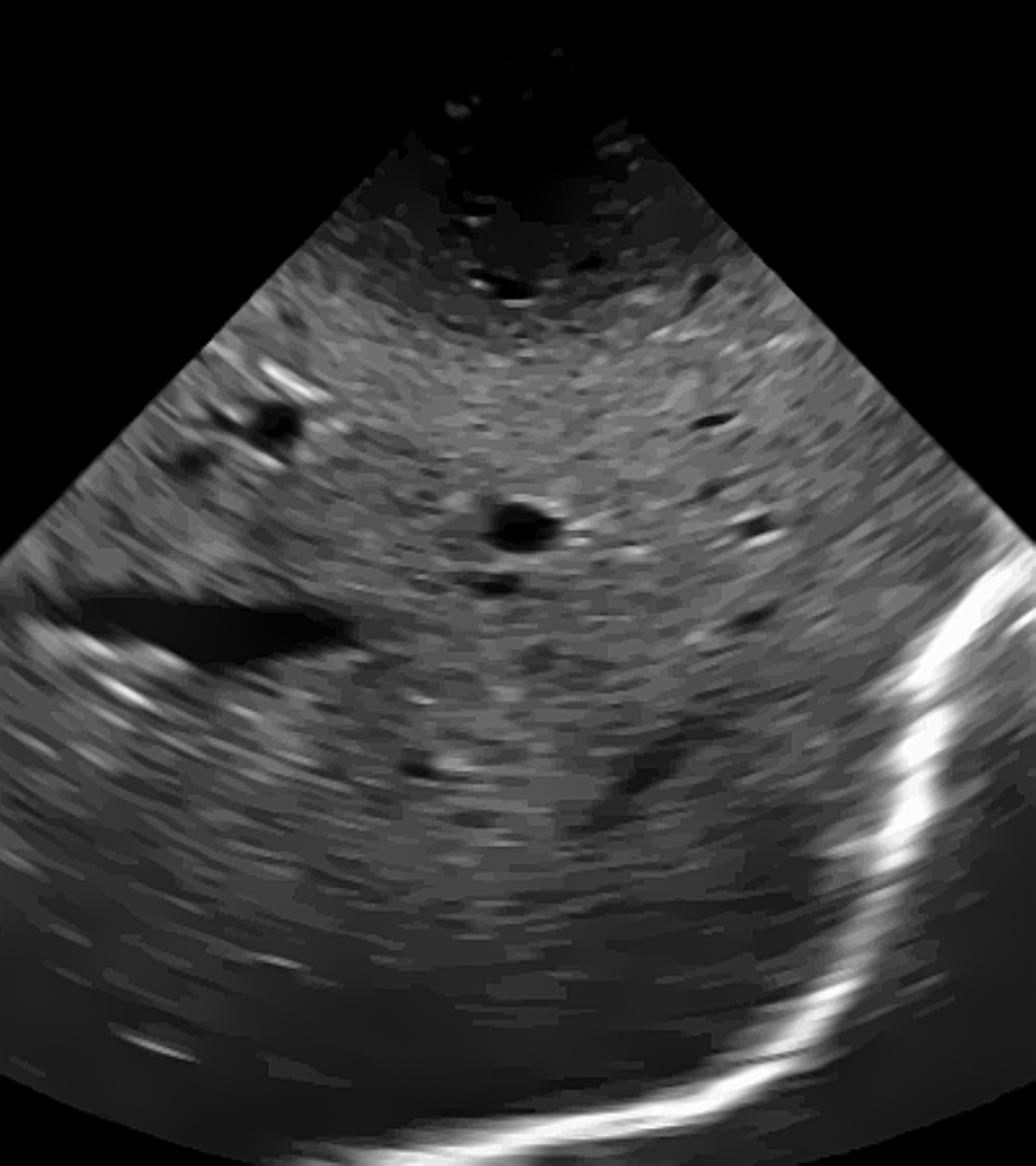}}
\end{minipage}
%
\vspace{10pt}

\begin{minipage}[]{0.23\linewidth}
  \centerline{\includegraphics[width=3.8cm]{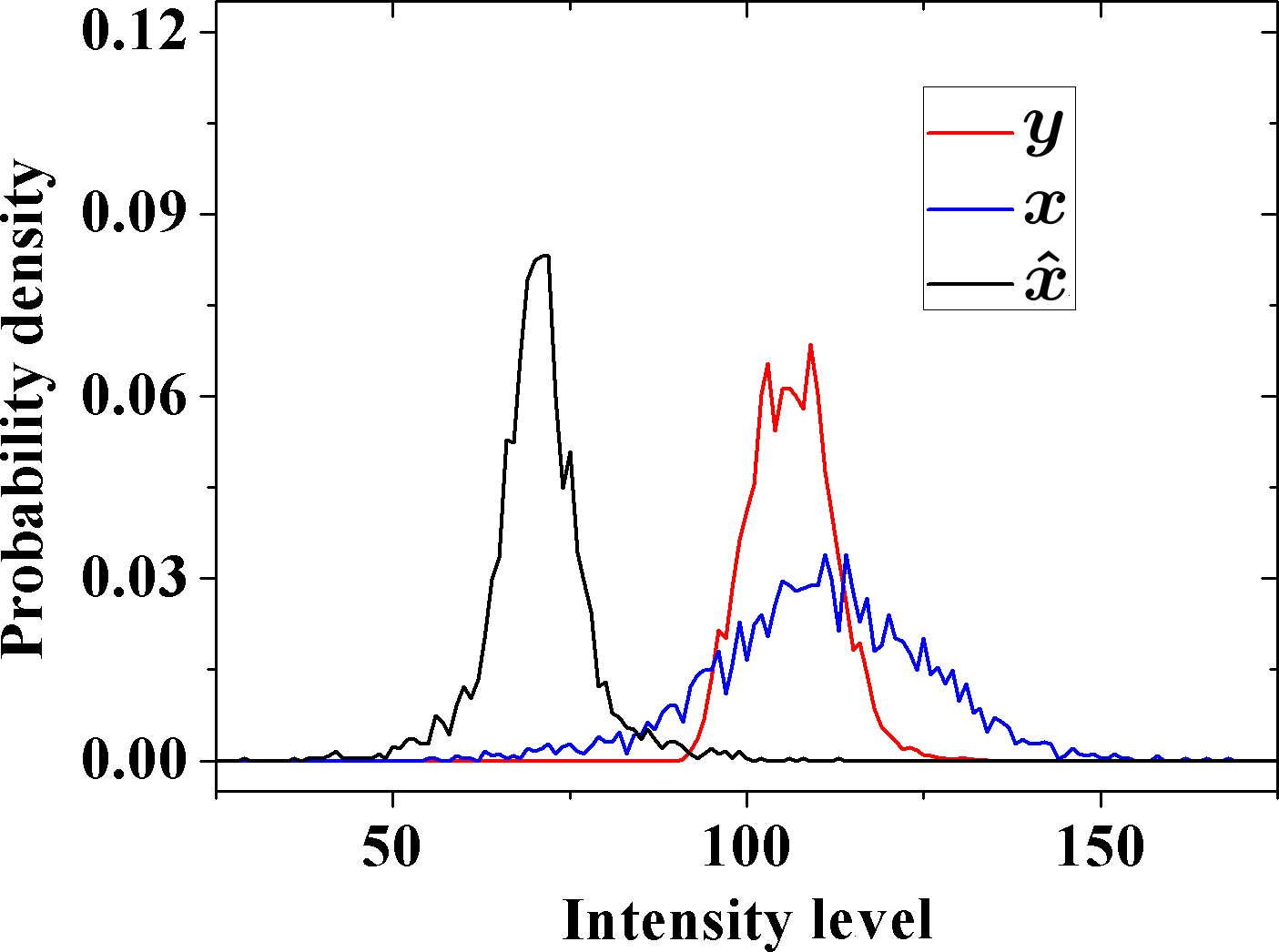}}
  \centerline{\footnotesize{(a) DPAD} }
\end{minipage}
%
\begin{minipage}[]{.23\linewidth}
  \centerline{\includegraphics[width=3.8cm]{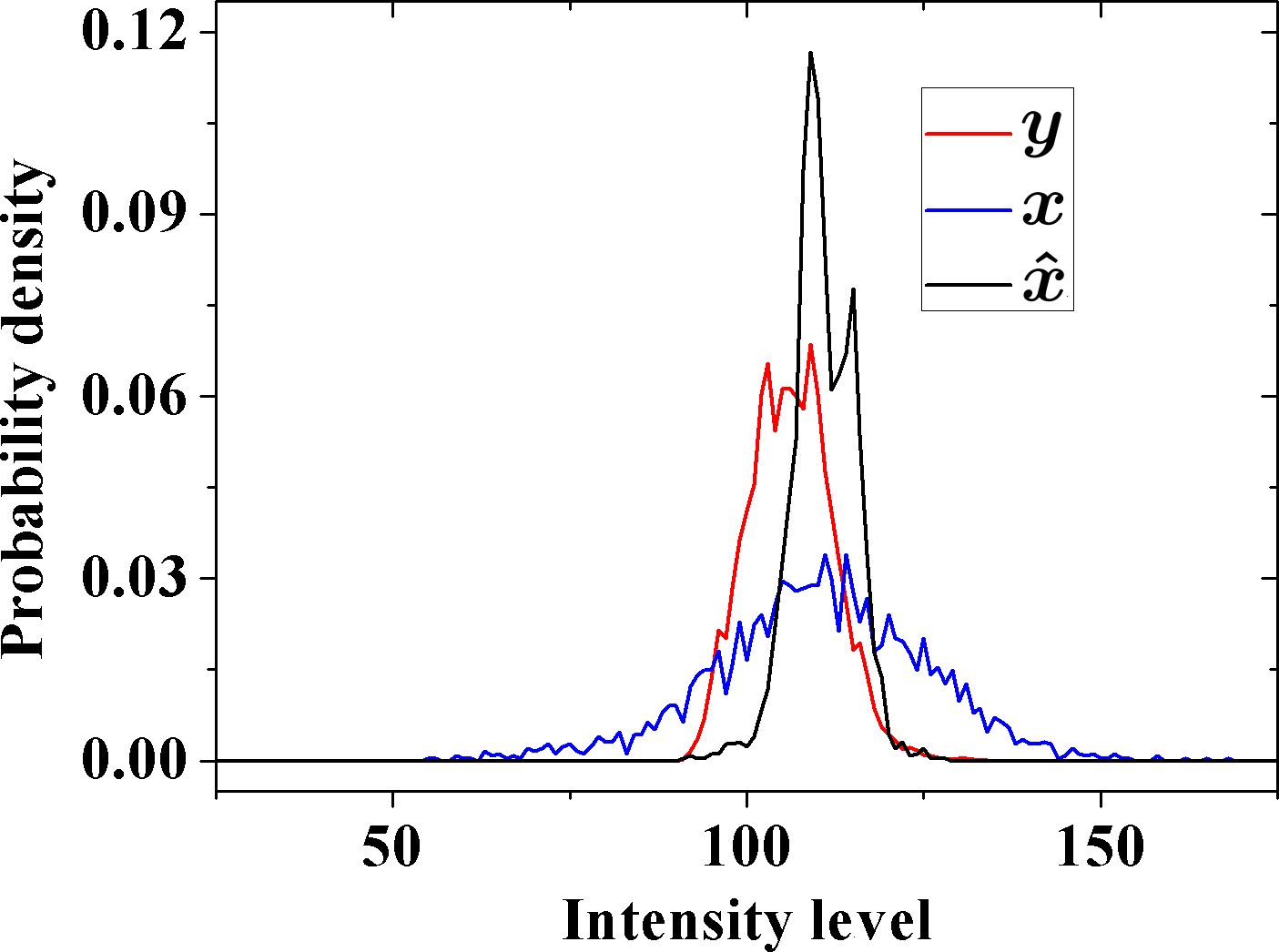}}
  \centerline{\footnotesize(b) OBNLM}
\end{minipage}
\begin{minipage}[]{0.23\linewidth}
  \centerline{\includegraphics[width=3.8cm]{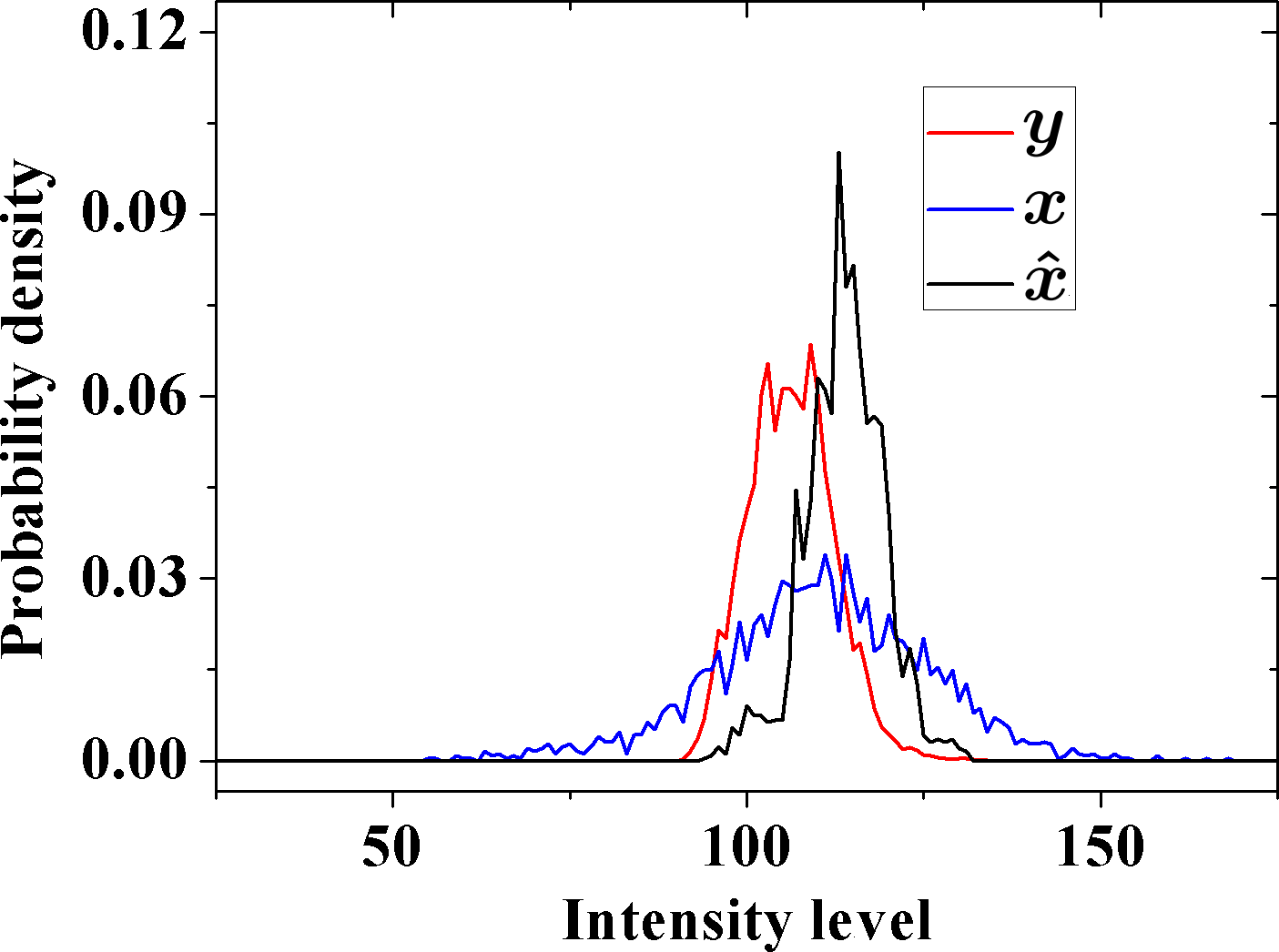}}
  \centerline{\footnotesize{(c) SBF} }
\end{minipage}
%
\begin{minipage}[]{.23\linewidth}
  \centerline{\includegraphics[width=3.8cm]{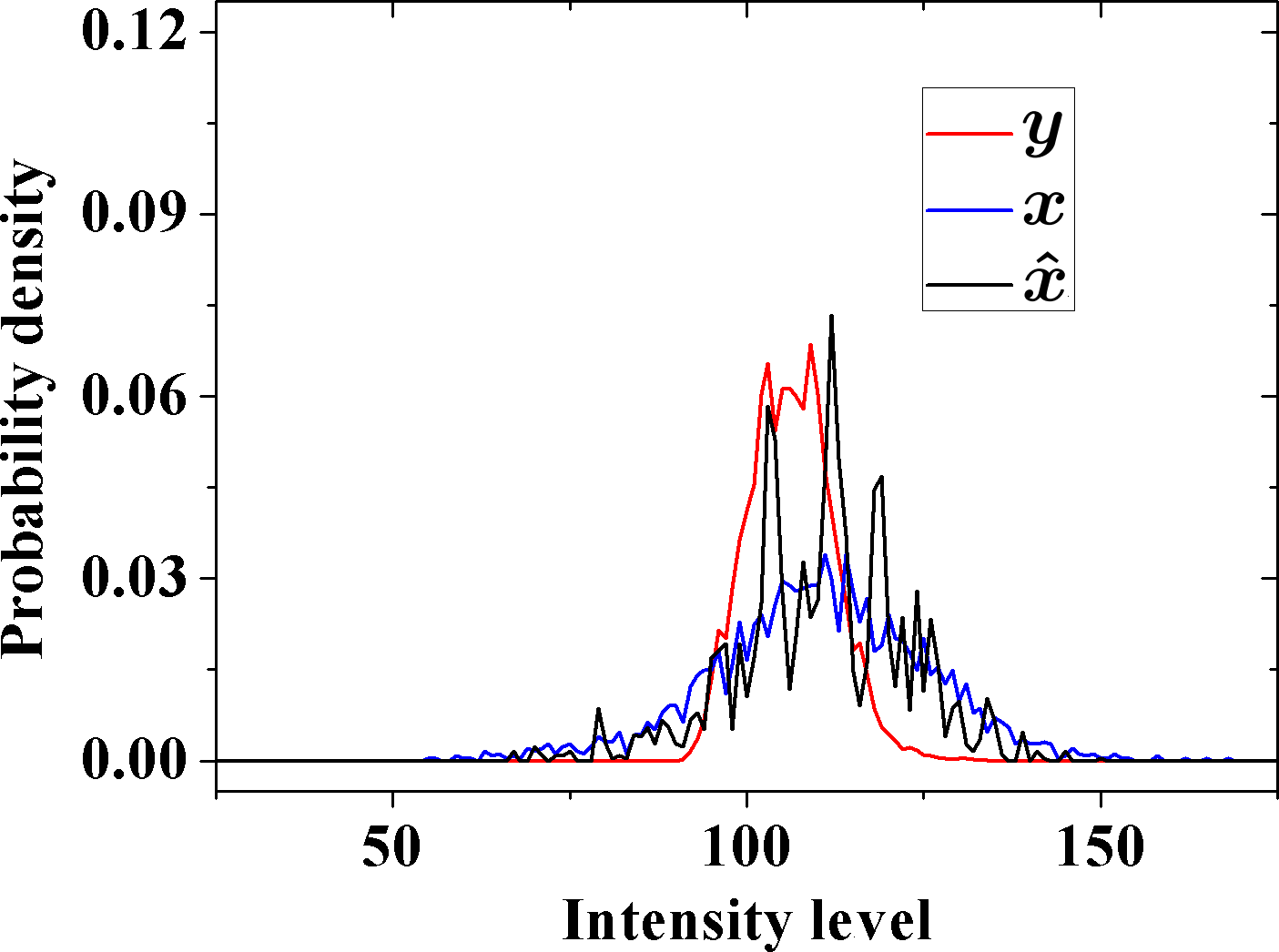}}
  \centerline{\footnotesize(d) RTAD}
\end{minipage}\\
\vspace{5pt}
\hspace{10pt}
\begin{minipage}[]{0.23\linewidth}
  \centerline{\includegraphics[width=3.8cm]{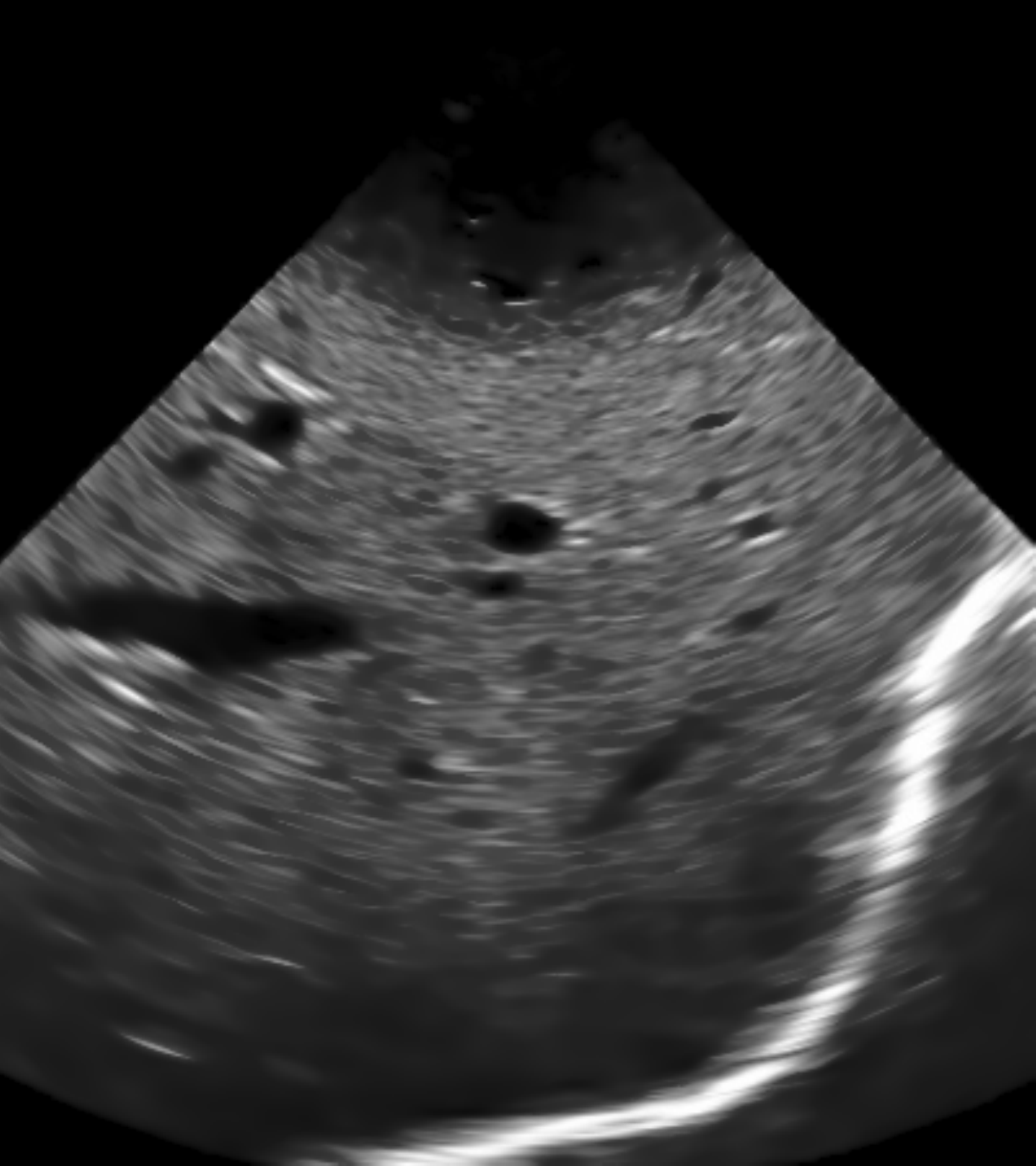}}
\end{minipage}
%
\begin{minipage}[]{0.23\linewidth}
  \centerline{\includegraphics[width=3.8cm]{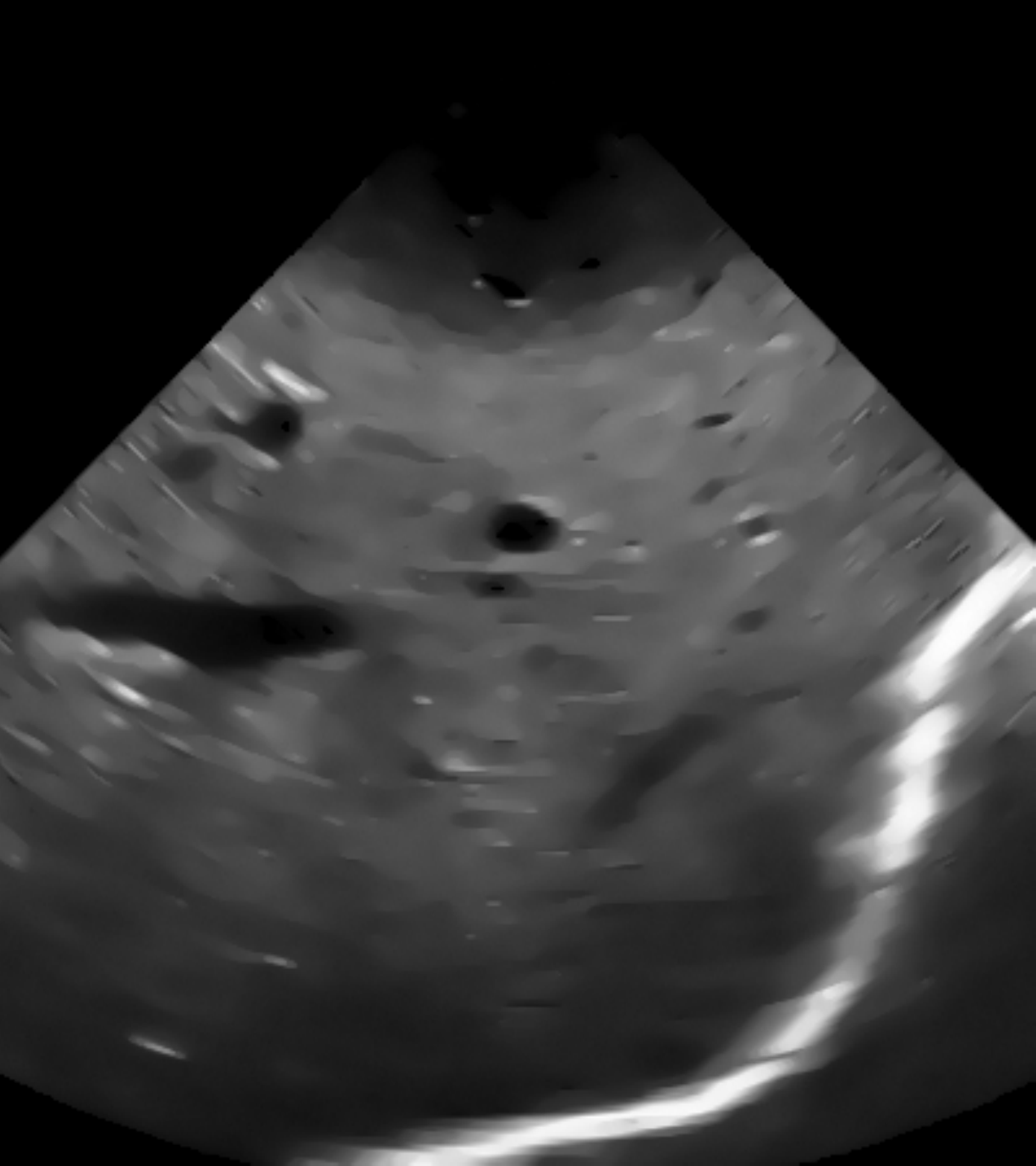}}
\end{minipage}
\begin{minipage}[]{0.23\linewidth}
  \centerline{\includegraphics[width=3.8cm]{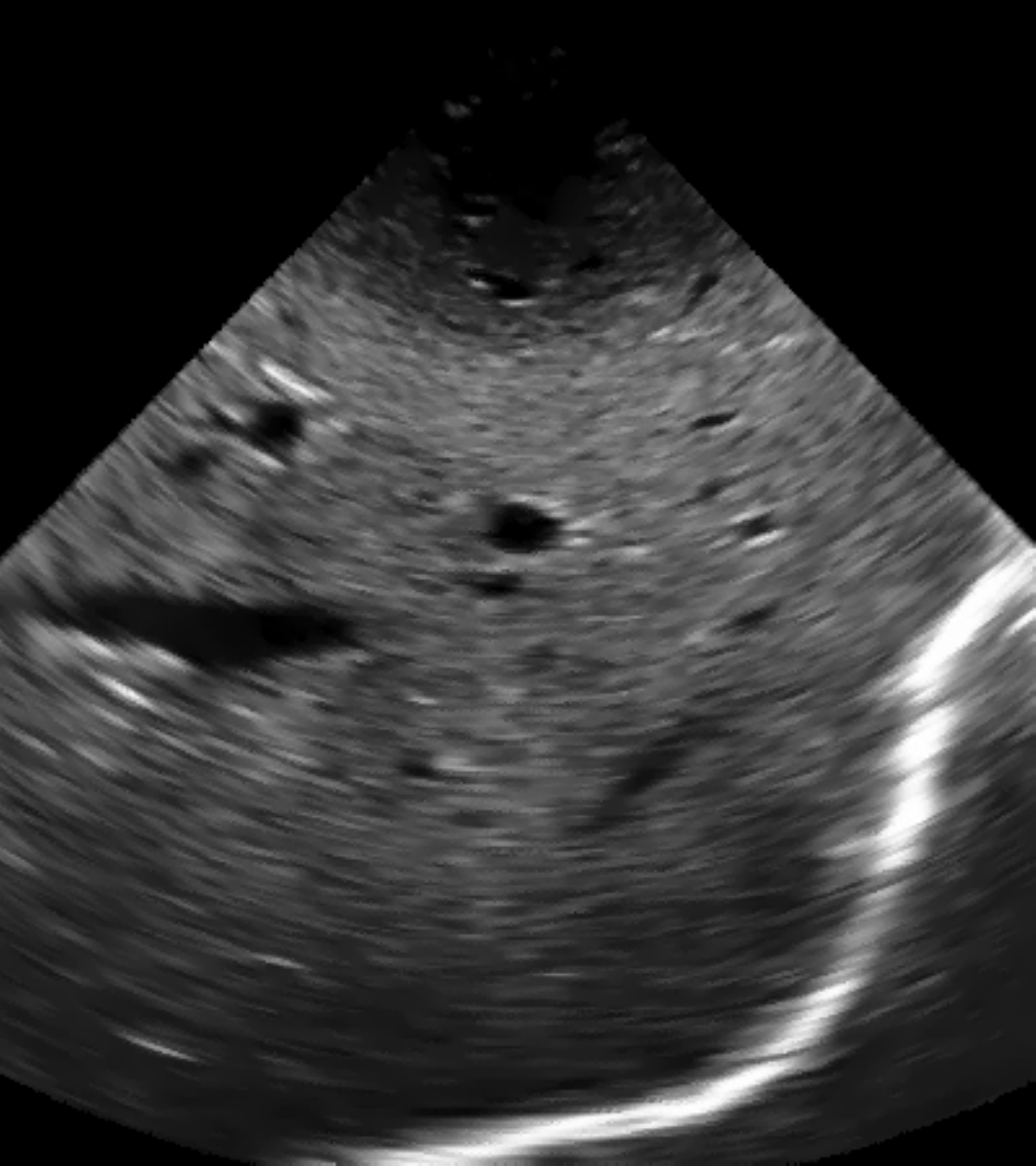}}
\end{minipage}
%
\begin{minipage}[]{.24\linewidth}
  \centerline{\includegraphics[width=3.8cm]{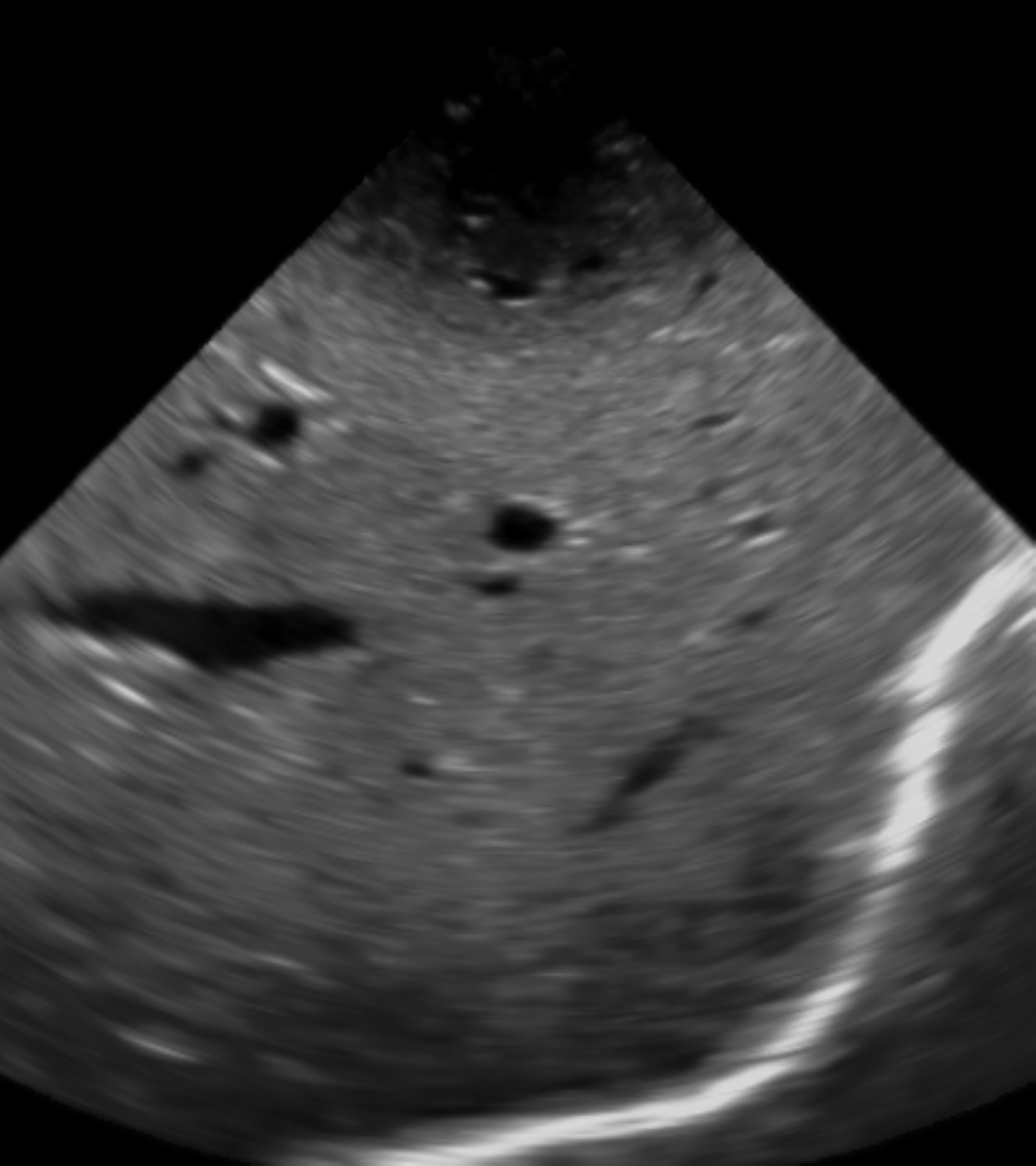}}
\end{minipage}
\hspace{-10pt}
\vspace{10pt}

\begin{minipage}[]{0.23\linewidth}
  \centerline{\includegraphics[width=3.8cm]{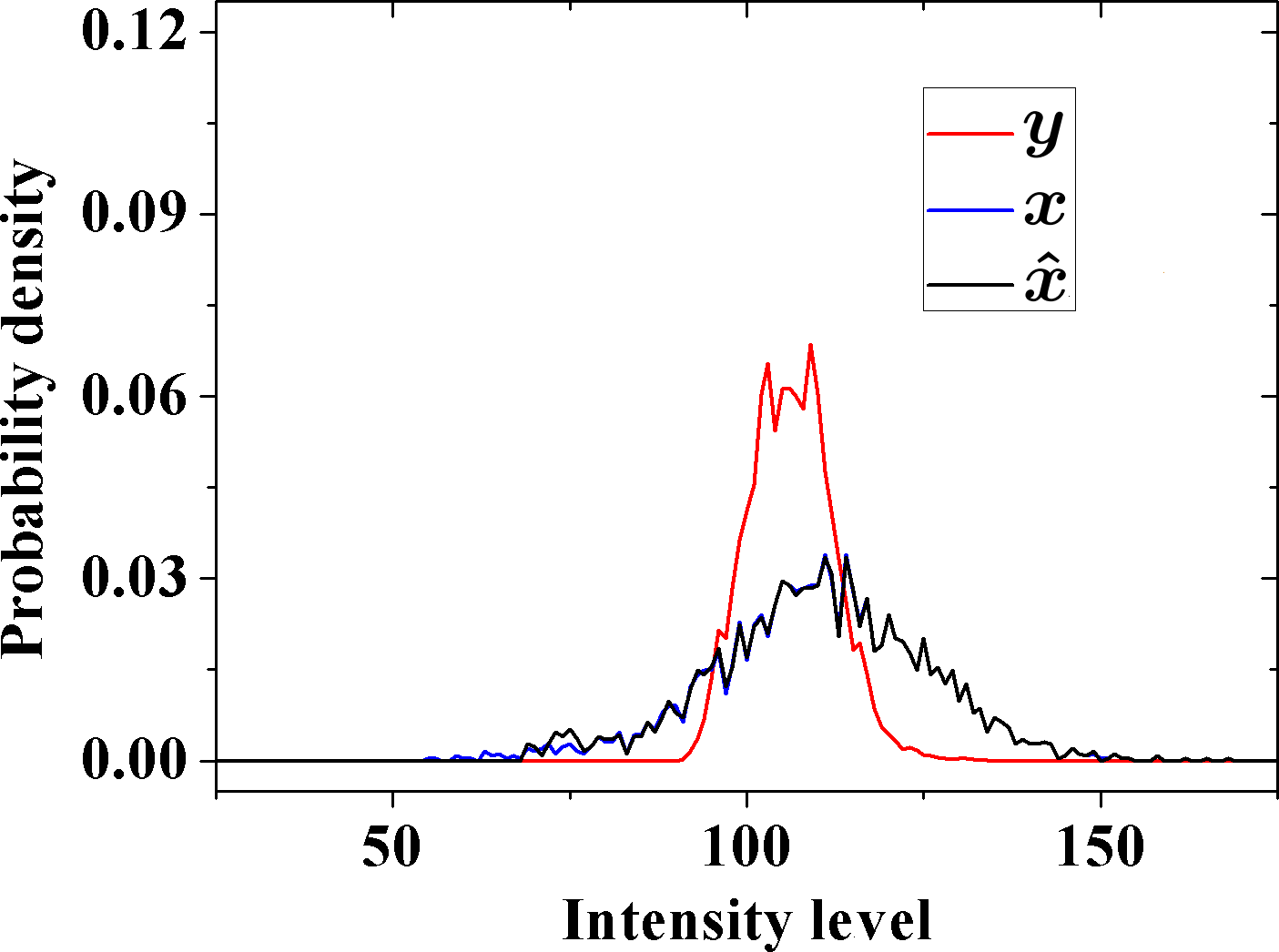}}
  \centerline{\footnotesize{(e) ADMSS} }
\end{minipage}
%
\begin{minipage}[]{0.23\linewidth}
  \centerline{\includegraphics[width=3.8cm]{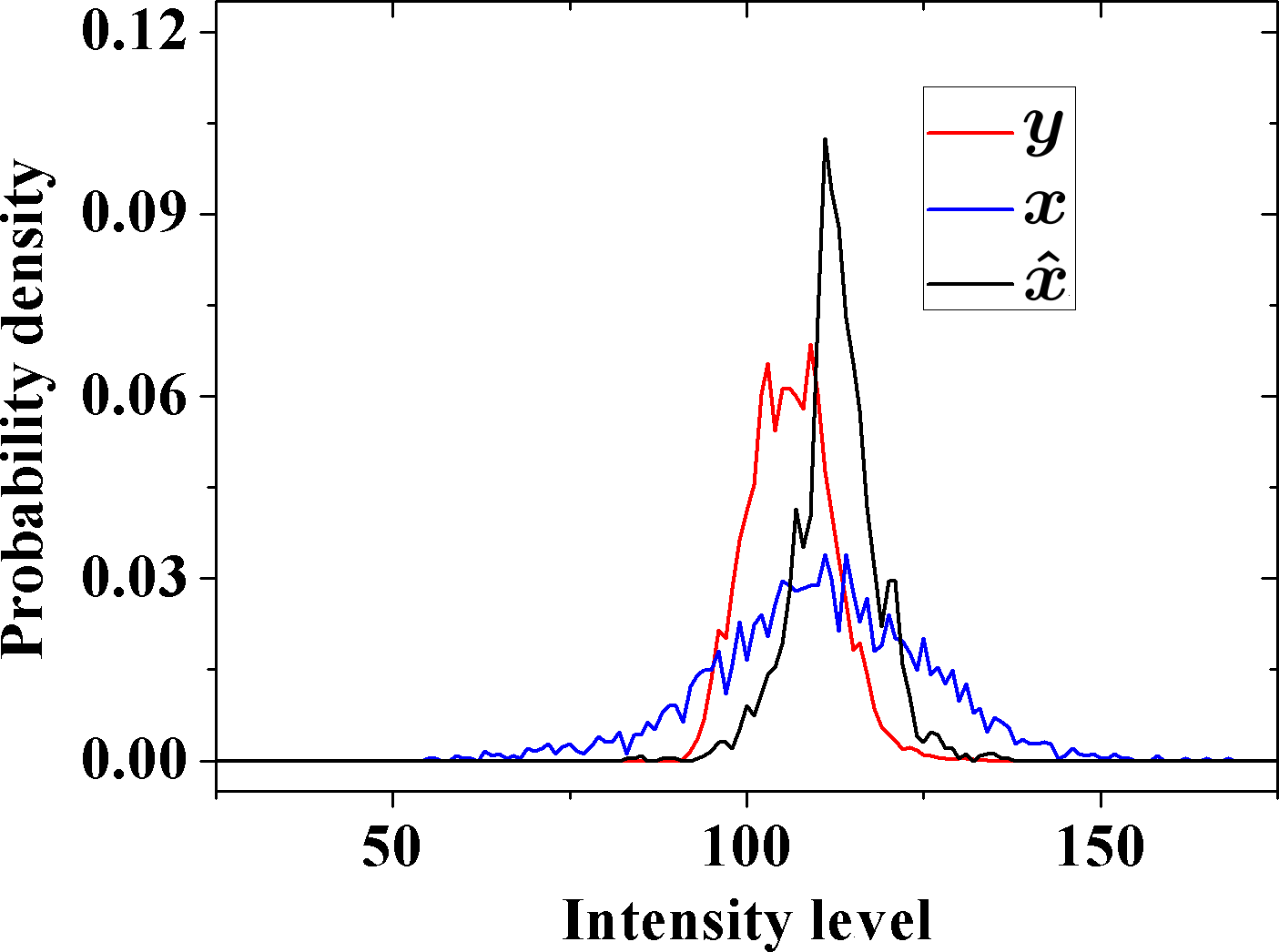}}
  \centerline{\footnotesize(f) EAGF}
\end{minipage}
\begin{minipage}[]{0.23\linewidth}
  \centerline{\includegraphics[width=3.8cm]{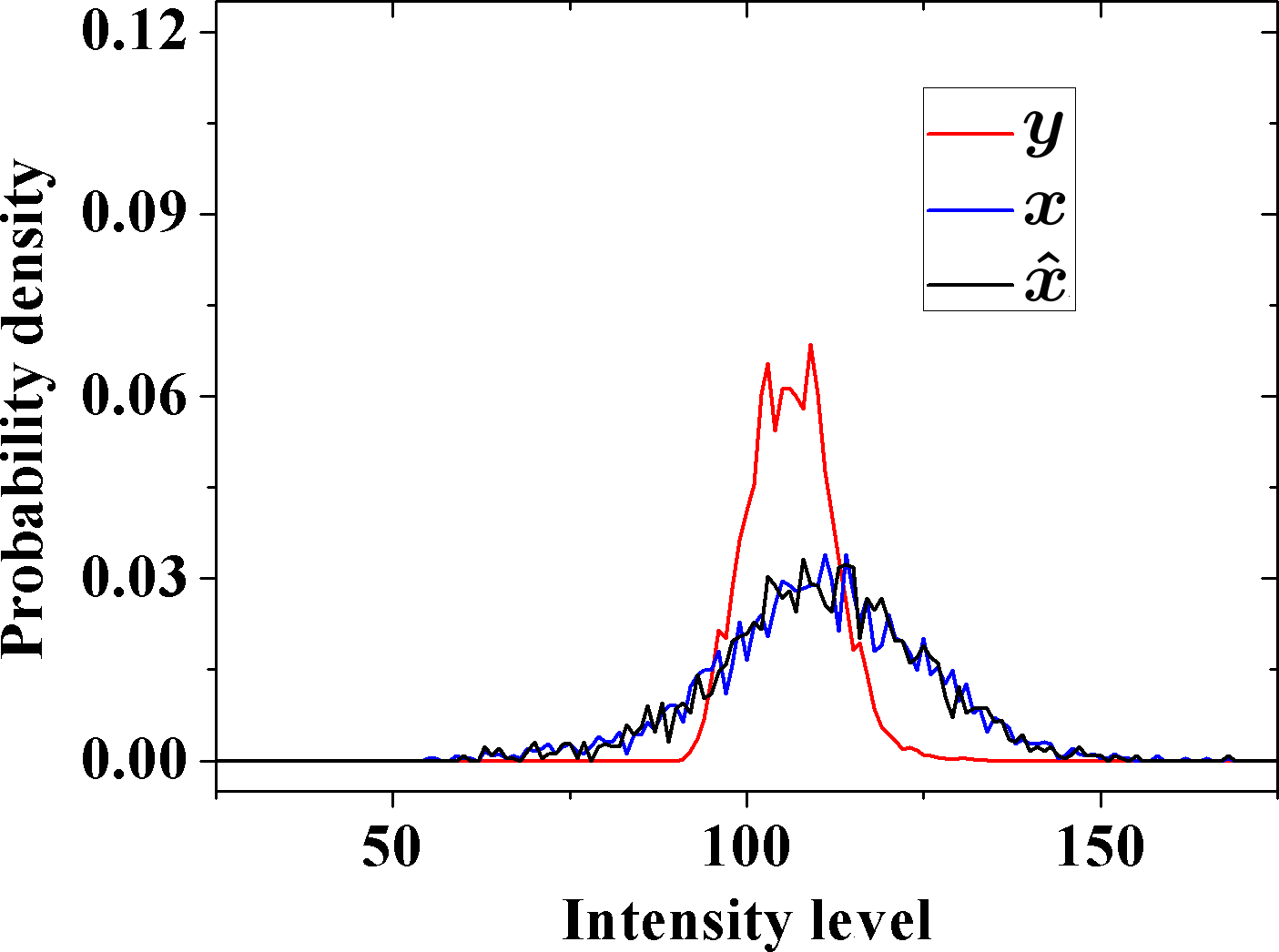}}
  \centerline{\footnotesize{(g) EPPRSRAD} }
\end{minipage}
%
\begin{minipage}[]{.24\linewidth}
  \centerline{\includegraphics[width=3.8cm]{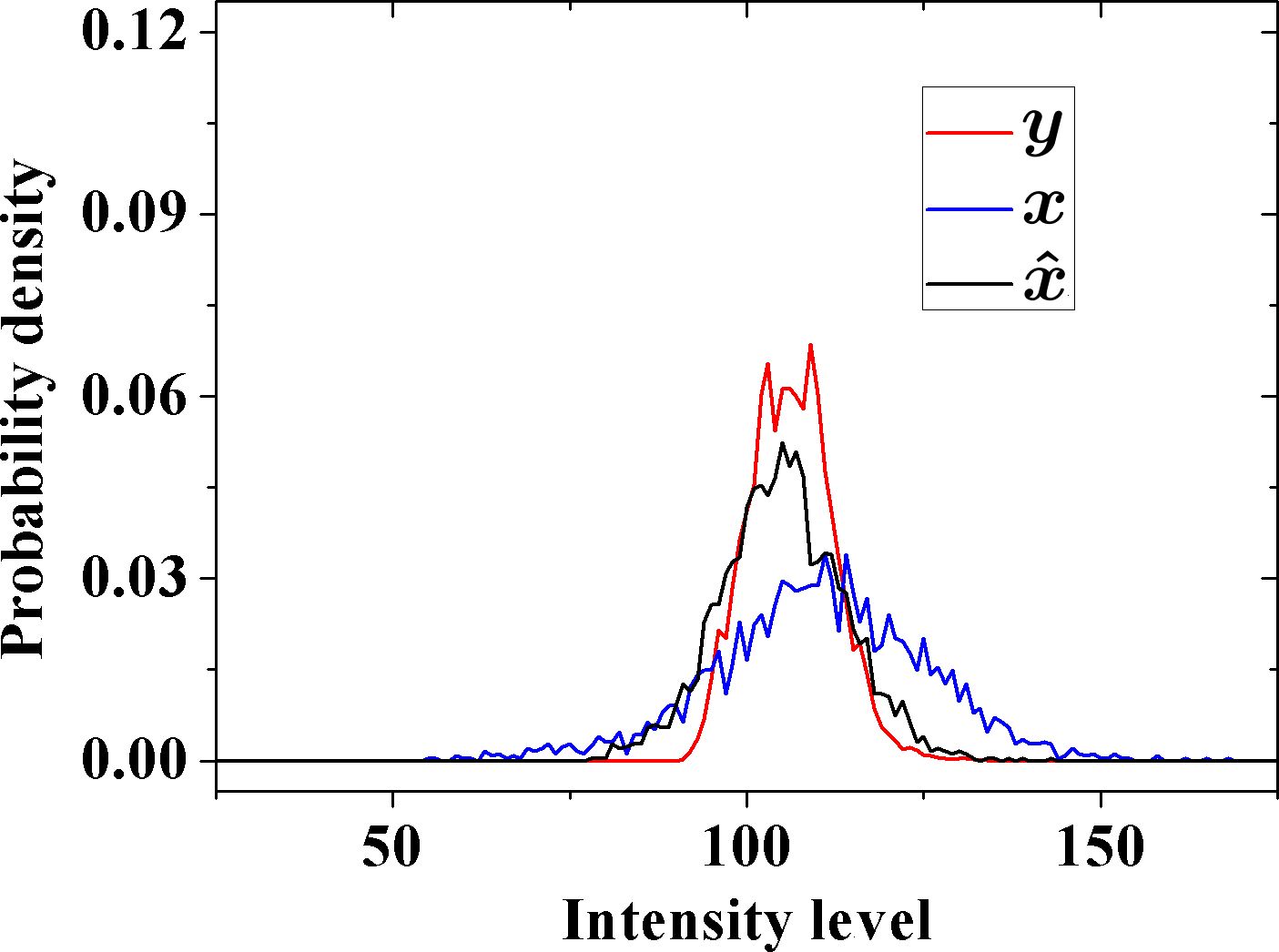}}
  \centerline{\footnotesize(h) Proposed}
\end{minipage}
\caption[]{Despeckled outputs obtained using different approaches. PDFs observed from the reference ($y$), input ($x$) and output ($\hat{x}$) are also shown. The proposed method results in the PDF closest to the reference.}
\label{out}
\end{figure*}

\subsection{Experimental results}
Performance of the proposed method is compared with DPAD \cite{aja2006estimation}, OBNLM \cite{coupe2009nonlocal}, SBF \cite{tay2010ultrasound}, RTAD \cite{deng2011speckle}, ADMSS \cite{ramos2015anisotropic}, EAGF \cite{mishra2017edge} and EPPRSRAD \cite{mishra2017edgeprob}. The implementation parameter values for DPAD, OBNLM, SBF and EAGF are obtained from \cite{mishra2017edge}. For the remaining methods, the values are taken from \cite{mishra2017edgeprob}. To compare the performance, two parenchymal regions of similar dimensions are identified in the images shown in Fig.~\ref{sample}. The probability density function (PDF) of the pixel intensities inside the marked regions are used to qualitatively compare the speckle characteristics. The PDFs obtained from the low speckle extent image ($y$, Fig.~\ref{sample}(a)), input image ($x$, Fig.~\ref{sample}(b)) and the despeckled images ($\hat{x}$) are shown in Fig.~\ref{out}. The despeckled outputs are also shown in Fig.~\ref{out}. 
The closest match to the desired PDF is produced by the proposed method. The observation is further validated by quantitative comparison of the obtained PDFs. KL-divergence is used for this purpose and the observed values are listed in Table~\ref{tab:1}. The smallest value of KL$_{({y}, {\hat{x}})}$ and a reasonable value of KL$_{({x}, {\hat{x}})}$, resulted in by the proposed method reflects its ability to produce the desired speckle characteristics. 

Further, the speckle extent in the identified regions is measured using the speckle region's signal to noise ratio (SSNR) \cite{mishra2017edgeprob}. The SSNR value observed from the reference image ($y$) is 18.2451. Nearest to this value is resulted in by EAGF, however, it is unable to provide the desired characteristics. In contrast, the proposed method gives optimal performance with a decent SSNR value and the desired speckle characteristics. Further, the proposed method is able to produce the despeckled output in less than 0.3 seconds on GPU and in about 16.5 seconds on CPU which is considerably smaller than the time taken by the approaches like OBNLM.

\begin{table}[!t]
	\centering
    \caption{{Quantitative evaluation of despeckling approaches}} \label{tab:1}
	\begin{tabular}[l]{|c|c|c|c|}
		\hline
		\bf Method & \bf SSNR$_{\boldsymbol{\hat{x}}}$ & \textbf{KL}$_{(\boldsymbol{x}, \boldsymbol{\hat{x}})}$ & \textbf{KL}$_{(\boldsymbol{y}, \boldsymbol{\hat{x}})}$ \\ \hline
		Input (no filter) & 7.2990 & 0 & 0.7512 \\ \hline
		DPAD & 9.7542 & 31.7018 & 31.5043\\ \hline
		OBNLM & 25.529 & 9.5010 & 0.9810\\ \hline
		SBF & 20.5514 & 8.8660 & 1.5440\\ \hline
		RTAD & 9.3614 & 1.9490 & 0.7325\\ \hline
		ADMSS & 7.3641 & 0.3242 & 0.7547\\ \hline
		EAGF & 19.3142 & 5.4180 & 1.0210\\ \hline
		EPPRSRAD & 7.5651 & 0.3428 & 0.7183\\ \hline
		Proposed & 12.016 & 3.0570 & 0.2107\\
		\hline
	\end{tabular}
\end{table}

\section{Conclusion}
In this paper, an unsupervised despeckling approach is presented.
The combination of adversarial and structural loss used to train the DRNN results in the despeckled images without losing the characteristic details. A better qualitative and quantitative performance of the proposed approach as compared to the state-of-the-art approaches is observed. The proposed work will provide motivation to explore the possibility of unsupervised deep learning for US image applications.


%



%

\ifCLASSOPTIONcaptionsoff
  \newpage
\fi



%
\newpage
\bibliographystyle{IEEEtran}
\bibliography{paper_gan}




%








\end{document}